# A Method of Generating Random Weights and Biases in Feedforward Neural Networks with Random Hidden Nodes

Grzegorz Dudek

*Abstract*—Neural networks with random hidden nodes have gained increasing interest from researchers and practical applications. This is due to their unique features such as very fast training and universal approximation property. In these networks the weights and biases of hidden nodes determining the nonlinear feature mapping are set randomly and are not learned. Appropriate selection of the intervals from which weights and biases are selected is extremely important. This topic has not yet been sufficiently explored in the literature. In this work a method of generating random weights and biases is proposed. This method generates the parameters of the hidden nodes in such a way that nonlinear fragments of the activation functions are located in the input space regions with data and can be used to construct the surface approximating a nonlinear target function. The weights and biases are dependent on the input data range and activation function type. The proposed methods allows us to control the generalization degree of the model. These all lead to improvement in approximation performance of the network. Several experiments show very promising results.

## I. Introduction

FEEDFORWARD neural networks (FNN) are extensively used in regression and classification applications due to their adaptive nature and universal approximation property. FNNs are able to learn from observed data and generalize well in unseen examples. The FNN inner parameters, i.e. weights and biases, are adjustable in the learning process. But due to the layered structure of the network this process is complicated, inefficient and requires the activation functions (AFs) of neurons to be differentiable. The training algorithms which involves the optimization of non-convex objective function, usually employ some form of gradient descent method which are known to be time consuming, sensitive to initial values of parameters and converging to local minima. Moreover some parameters, such as number of hidden nodes or learning algorithm parameters, have to be tuned manually.

In recent years, alternative learning methods have been developed, in which the network parameters are selected randomly, so that the resulting optimization task becomes convex and can be formulated as a linear least-squares problem [1]. Such methods are applied in three broad families of NNs: FNNs, recurrent NNs, and randomized kernel approximations [2]. Many simulation studies reported in the literature show high performance of the randomized models which is compared to fully adaptable ones. Randomization which is cheaper than optimization provides to simplicity in implementation and faster training.

In feedforward neural networks with random hidden nodes (FNNRHN), the learning process does not require iterative tuning of weights. The weights and biases of hidden neurons need not to be adjusted. They are randomly selected from some intervals according to any continuous sampling distribution and remain fixed. The only parameters need to be learned are the output weights, linking the hidden and output nodes. Thus, FNNRHN can be considered as a linear system in which the output weights are analytically determined through simple generalized inverse operation of the hidden layer output matrices. For this reason, the learning speed can be thousands of times faster than classical gradient descent-based learning. As theoretical studies have shown [3], when the parameters of the hidden nodes are randomly generated from a uniform distribution within a proper range, the resulting neural network is a universal approximator for a continuous function on a bounded finite dimensional set with efficient convergence rate. Husmeier in [4] proved that the universal approximation property also holds for symmetric interval setting of the random parameter scope if the function to be approximated meets Lipschitz condition. However, how to select the range for the random parameters remains an open question. This issue is considered to be one of the most important research gaps in the field of randomized algorithms for training NNs [5]. These algorithms suffer from design choices, translated in free parameters, which are difficult to set optimally and require many trials and cross-validation to find a good projection space [1].

The authors of new solutions in NNs with randomization do not give any hints on the ranges for random parameters [6], [7]. So, in the applications of FNNRHNs to classification or regression problems the ranges for random parameters of hidden nodes are selected without scientific justification and could not ensure the universal approximation property of the network. Usually these intervals are assigned as fixed (typically $[-1, 1]$ for weights and $[0, 1]$ for biases), regardless of the

The author is with the Department of Electrical Engineering, Czestochowa University of Technology, 42-200 Czestochowa, Al. Armii Krajowej 17, Poland (e-mail: dudek@el.pcz.czest.pl).

data and the AF type. In some papers independency of hidden neurons on data is seen as an asset [8].

The problem with selection of appropriate ranges for random parameters of the hidden nodes is not solved till today. However, in many works concerning FNNRHNs attention is drawn to the significance of the intervals from which random weights and biases are selected. In conclusion of [9] it is rightly pointed out that when network nodes are chosen at random and not subsequently trained, they are usually not placed in accordance with the density of the input data. In such a case training of linear parameters becomes ineffective at reducing errors. Moreover, the number of nodes needed to approximate a nonlinear map grows exponentially, and the model is very sensitive to the random parameters. To improve effectiveness of the network the authors of [9] advice combining unsupervised placement of network nodes according to the input data density with subsequent supervised or reinforcement learning values of the linear parameters of the approximator. This work motivated the authors of [10] to highlight some risky aspects caused by the randomness in FNNRHN, such as the illogical way of simply selecting a trivial range [−1, 1] for random assignment of the input weights and biases. They analyze some impacts of the scope of random parameters on the model performance, and empirically show that a widely used setting for this scope is misleading. Although, they observe that for some specific scopes the network performs better in both learning and generalization than in other scopes, they do not give tips on how to select appropriate scopes. There is no such tips also in [11], where authors investigate the range for random parameters by introducing a scaling factor to control this range. The work is concluded that scaling down the randomization range to avoid saturating the neurons may risk at degenerating the discrimination power of the random features. Scaling this range up to enhance the discrimination power of the random features may risk saturating the neurons.

In some papers we can find some suggestions on how to generate random parameters of hidden neurons. In the early work on FNNs with randomization [12], the parameters of hidden nodes were set to be uniform random values in [−1, 1], but authors suggest to optimize this range in a more appropriate range for the specified application. In [4] the author suggests to use symmetric and "large enough" boundaries for the hidden node parameters and advices to optimize them in the training process. In [13] a valuable remark on the selection of hidden node parameters appears: they should be generated at random but then scaled to avoid saturation of AF. Unfortunately, any further tips on scaling are not given. More details on generating random parameters of hidden nodes in [14] are given. The weights are chosen from a normal distribution with zero mean and some specified variance that can be adjusted to obtain input-to-node values that do not saturate the sigmoids. The biases are computed to center each sigmoid at one of the training points. This distributes the sigmoids across the input space, as is suggested by the Nguyen–Widrow weight initialization algorithm [15].

Lately, Wang and Li proposed a supervisory mechanism of assigning the input weights and biases of the hidden nodes in their learner model generated incrementally by stochastic configuration algorithms [16]. The random parameters are generated with an inequality constraint adaptively selecting the scope for them, ensuring the universal approximation property of the model. The authors adopt from [4] the symmetric interval setting for the random parameters. The scope $[-\lambda, \lambda]$ is searched in the iterative procedure, from $\lambda = \lambda_{\min}$ to $\lambda = \lambda_{\max}$. The input weights and biases are generated both from the same symmetric interval. This is against the conclusion of this work: the input weights and biases should be generated from different ranges because they have different meaning. A method proposed in this work generate them separately depending on the data (its scope and complexity) and activation function type. In the experimental part of the work we compare the results of the proposed method and the stochastic configuration network [16].

In our previous works [17], [18] we looked inside FNNRHN and studied how the AFs of neurons compose the fitting curve and how the ranges from which weights and biases are randomly generated affect the approximation ability of the network. The aim of those works was to provide a guidance for how to randomly generate the weights and biases to get good performance in approximation of the functions of one variable. In this work we focus on multidimensional cases. A method of randomly generating FNNRHN parameters to set nonlinear fragments of AFs in the input space regions containing data points is proposed. This method allows us to control the flatness and steepness of AFs in the input hypercube and hence the degree of generalization of the network.

The remainder of the paper is organized as follows. Section II briefly presents FNNRHN learning algorithm. In Section III the intervals for random parameters are determined on the basis of theoretical analysis for one-dimensional case. The analysis were performed for four popular AFs: sigmoidal, Gaussian, softplus and sine/cosine. Similar analysis were performed for multidimensional case in Section IV. Section V reports the simulation study and compare results of the proposed method with the newest results from the literature. Section VI concludes the paper.

## II. FNNRHN LEARNING ALGORITHM

The architecture of FNNRHN is the same as for single-hidden-layer feedforward neural network. One output is considered, $m$ hidden neurons and $n$ inputs. The training set is $\Phi = \{(\mathbf{x}_l, y_l) \mid \mathbf{x}_l \in \mathrm{R}^n, y_l \in \mathrm{R}, l = 1, 2, \ldots, N\}$ and the AF of hidden nodes is $h(\mathbf{x})$. The learning algorithm consists of three steps.

1. Randomly generate hidden node parameters: weights $\mathbf{a}_i = [a_{i,1}, a_{i,2}, \ldots, a_{i,n}]^\mathrm{T}$ and biases $b_i$, $i = 1, 2, \ldots, m$, according to any continuous sampling distribution. Usually $a_{i,j} \sim \mathrm{U}(a_{\min}, a_{\max})$ and $b_i \sim \mathrm{U}(b_{\min}, b_{\max})$.
2. Calculate the hidden layer output matrix $\mathbf{H}$:

$$\mathbf{H} = \begin{bmatrix} \mathbf{h}(\mathbf{x}_1) \\ \vdots \\ \mathbf{h}(\mathbf{x}_N) \end{bmatrix} = \begin{bmatrix} h_1(\mathbf{x}_1) & \cdots & h_m(\mathbf{x}_1) \\ \vdots & \vdots & \vdots \\ h_1(\mathbf{x}_N) & \ldots & h_m(\mathbf{x}_N) \end{bmatrix} \qquad (1)$$

where $h_i(\mathbf{x})$ is an AF of the *i*-th node, which is nonlinear piecewise continuous function, e.g. a sigmoid:

$$h_i(\mathbf{x}) = \frac{1}{1+\exp(-(\mathbf{a}_i^T\mathbf{x}+b_i))} \quad (2)$$

The *i*-th column of $\mathbf{H}$ is the *i*-th hidden node output vector with respect to inputs $\mathbf{x}_1, \mathbf{x}_2, ..., \mathbf{x}_N$. Hidden neurons map the data from *n*-dimensional input space to *m*-dimensional feature space, and thus, $\mathbf{h}(\mathbf{x}) = [h_1(\mathbf{x}), h_2(\mathbf{x}), ..., h_m(\mathbf{x})]$ is a nonlinear feature mapping. The output matrix $\mathbf{H}$ remains unchanged because parameters of the AFs, $\mathbf{a}_i$ and $b_i$, are fixed.

3. Calculate the output weights $\beta_i$:

$$\boldsymbol{\beta} = \mathbf{H}^+\mathbf{Y} \quad (3)$$

where $\boldsymbol{\beta} = [\beta_1, \beta_2, ..., \beta_m]^T$ is a vector of output weights, $\mathbf{Y} = [y_1, y_2, ..., y_N]^T$ is a vector of target outputs, and $\mathbf{H}^+$ is the Moore–Penrose generalized inverse of matrix $\mathbf{H}$.

The above equation for $\boldsymbol{\beta}$ results from the minimizing the approximation error:

$$\min \|\mathbf{H}\boldsymbol{\beta} - \mathbf{Y}\|^2 \quad (4)$$

The function expressed by FNN is a linear combination of the AFs $h_i(\mathbf{x})$. In the one output case it is of the form:

$$\varphi(\mathbf{x}) = \sum_{i=1}^m f_i(\mathbf{x}) = \sum_{i=1}^m \beta_i h_i(\mathbf{x}) = \mathbf{h}(\mathbf{x})\boldsymbol{\beta} \quad (5)$$

where $f_i(\mathbf{x}) = \beta_i h_i(\mathbf{x})$ is the weighted output of the *i*-th hidden node.

The presented network is the most popular solution of FNNRHN. But it should be mentioned, that the prototype of NN with randomization, i.e. Random Vector Functional Link (RVFL) network proposed by Pao and Takefji [19], has direct links from the input layer to the output one.

## III. GENERATING RANDOM WEIGHTS AND BIASES - ONE-DIMENSIONAL CASE

For brevity, we use the following acronyms:
- TF: target function $g(x)$,
- FC: fitted curve $\varphi(x)$,
- II: input interval, i.e. the interval to which inputs are normalized.

To illustrate results the single-variable TF is used of the form:

$$g(x) = \sin(20\cdot\exp(x))\cdot x^2 \quad (6)$$

where $x \in [0, 1]$.
A variation of this function increases along the II [0, 1] (see top chart in Fig. 2). At the left border of the II it is flat, while towards the right border it expresses increasing oscillations.

The training set contains 5000 points $(x_l, y_l)$, where $x_l$ are uniformly randomly distributed on [0, 1] and $y_l$ are distorted by adding the uniform noise distributed in [–0.2, 0.2]. The testing set of the same size is created similarly but without noise. The outputs are normalized into the range [–1, 1].

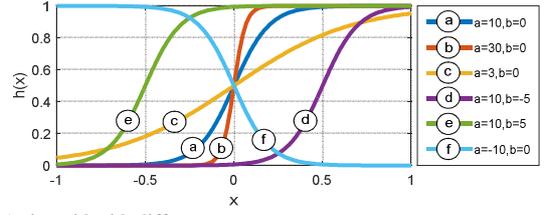

Fig. 1. A sigmoid with different parameters.

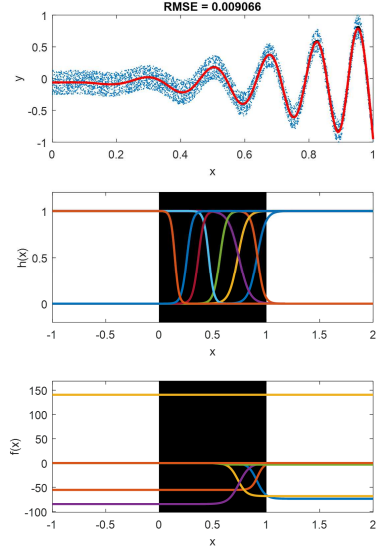

Fig. 2. Results of fitting for FNN with 9 hidden sigmoid nodes learned using Levenberg-Marquardt algorithm.

*A. Sigmoid AFs*

Let us look inside NN and analyze how the FC is constructed. Let a sigmoid be an AF of hidden neurons:

$$h(x) = \frac{1}{1+\exp(-(a\cdot x+b))} \quad (7)$$

The weight *a* decides about a slope of the sigmoid and the bias *b* shifts the function along the x-axis (see Fig. 1). For positive *a* the slope of the sigmoid ($dh/dx$) is positive, and for negative *a* the slope is negative. The set of hidden neurons represents a set of AFs which are combined linearly to produce FC.

Results of curve (6) fitting when using single-hidden layer FNN with 9 hidden neurons in Fig. 2 are shown. FNN was trained using Levenberg-Marquardt algorithm [20]. The middle chart shows AFs of 9 hidden neurons with parameters $a_i$ and $b_i$. These parameters are determined in the learning process, as well as the output weights $\beta_i$. The bottom chart shows AFs multiplied by the output weights $\beta_i$. The sum of these curves gives FC, which is drawn with a solid line in the upper chart. Note that AFs have their nonlinear, steep fragments inside the II (shown in black). These fragments are used to compose a FC. When the TF expresses complex behavior, such as function (6), AFs should be distributed in the II in such a way that their steep fragments correspond to the steep fragments of the TF.

Now, let us use FNNRHN with 100 hidden neurons for fitting curve (6). Let us generate randomly weights and biases over the interval [−1, 1], which is typical for FNNRHN [12]. As we can see from Fig. 3 the AF fragments in the II are too flat and cannot be combined to get our TF. Another example in Fig. 4 is shown. Here weights are generated from [−10, 10] and biases from [−1, 1]. As we can see from this figure the steep fragments of AFs are at the left border, where the TF is flat. On the other hand, at the right border, where the TF requires steep fragments, there are the flat AF fragments. This results in poor fitting. The above examples show that the problem is in definition of appropriate intervals for random weights and biases.

To determine the interval for $a$, let us set a sigmoid $S$ in the II in such a way that its inflection point (which is for $h(x) = 0.5$) is in $x = 0$ and the sigmoid value in $x = 1$ is $r \in (0, 0.5)$ (see top, left chart in Fig. 5). Note that in such case the most nonlinear and steepest fragment of a sigmoid, which is around the inflection point, is inside the II. The parameter $r$ should be lower than 0.5 (sigmoid value for the inflection point). For $r = 0.5$ we have completely flat function. If $r$ decreases toward 0, the sigmoid $S$ is more and more steep in the II. Thus $r$ controls the flatness of $S$ in the II.

When we set the inflection point of $S$ in $x = 0$, and the sigmoid value for $x = 1$ is $r$, then the shift parameter $b = 0$ and:

$$\frac{1}{1+\exp(-(a \cdot 1 + 0))} = r \quad (8)$$

After transformations we get from (8) a slope parameter for $S$:

$$a = -\ln\left(\frac{1-r}{r}\right) = a_{\lim 1} \quad (9)$$

(For $r \in (0, 0.5)$ $a_{\lim 1}$ is negative.)

Let us assume that the AFs building the FC are not flatter than the sigmoid $S$. Thus, their slope parameters satisfy the condition:

$$a \leq -|a_{\lim 1}| \quad \text{or} \quad a \geq |a_{\lim 1}| \quad (10)$$

Parameter $a_{\lim 1}$ defines the flattest AF possible in the set of $m$ AFs. Let:

$$a_{\lim 2} = s \cdot a_{\lim 1} \quad (11)$$

where $s > 1$ defines the steepest AF possible. So, the slope parameter of the $i$-th AF can be generated from the ranges:

$$a_i \in [-|a_{\lim 2}|, -|a_{\lim 1}|] \cup [|a_{\lim 1}|, |a_{\lim 2}|] \quad (12)$$

After substituting (9) and (11) in (12) and simplifying notation we obtain:

$$|a_i| \in \left[\ln\left(\frac{1-r}{r}\right), s \cdot \ln\left(\frac{1-r}{r}\right)\right] \quad (13)$$

Parameter $s$ decides about the maximal steepness of AFs, and should correspond to the steepness of the TF.

When AF satisfies condition (13) it lies between two boundary AFs, with slope parameters $a_{\lim 1}$ and $a_{\lim 2}$. These boundary AFs allow us to control the steepness of hidden neuron AFs to avoid their saturation fragments in the input interval. This is because saturation fragments are not suitable for nonlinear function fitting.

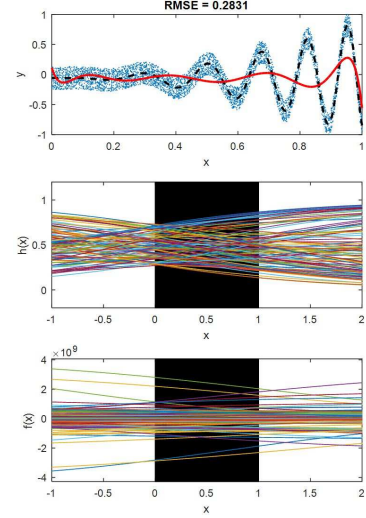

Fig. 3. Results of fitting for FNNRHN with 100 hidden sigmoid nodes, $a, b \in [-1, 1]$.

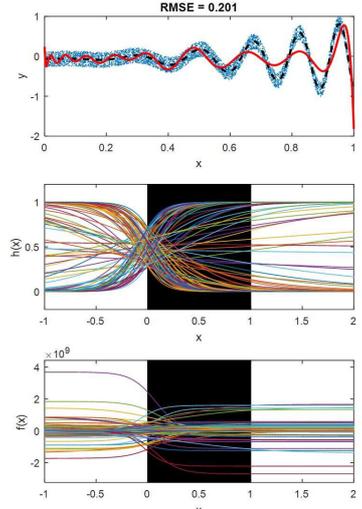

Fig. 4. Results of fitting for FNNRHN with 100 hidden sigmoid nodes, $a \in [-10, 10]$ and $b \in [-1, 1]$.

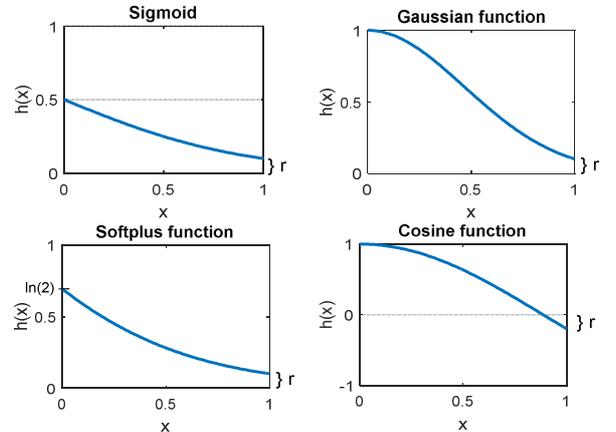

Fig. 5. The flattest fragments of AFs in the II.

Now, let us set the shift parameter $b$ in such a way that the sigmoid inflection point is inside the II. So, for some $x \in [0, 1]$ we get:

$$\frac{1}{1+\exp(-(a \cdot x + b))} = 0.5 \qquad (14)$$

After transformations we obtain:

$$b = -a \cdot x \qquad (15)$$

For $x = 0$ we get a border of the interval for $b$: $b_{\lim 1} = 0$, and for $x = 1$, we get the second border of this interval: $b_{\lim 2} = -a$. Note, that the interval for the shift parameter $b$ of some AF is dependent on the value of the slope parameter $a$ of this AF. Thus, the biases should be generated individually for each $i$-th AF from the interval:

$$b_i \in \begin{cases} [0, a_i] & \text{for } a_i \leq 0 \\ [-a_i, 0] & \text{for } a_i > 0 \end{cases} \qquad (16)$$

In Fig. 6 results of fitting are shown, where FNNRHN has 100 hidden nodes and the above described approach is used for generating random weights and biases. For $r = 0.1$ and $s = 3$ from (13) we get: $|a_i| \in [2.20, 6.56]$. Too small value of $s$ leads to underfitting and too high value leads to overfitting (see. Fig. 7). So it is recommended to select this parameter experimentally, e.g. in the cross-validation procedure, as well as parameter $r$.

B. *Gaussian AFs*

Now, let us consider Gaussian AF of the form:

$$h(x) = \exp(-(a \cdot x + b)^2) \qquad (17)$$

Similarly to a sigmoid, weight $a$ decides about a slope or width of the Gaussian function and the bias $b$ shifts the function along the x-axis. To determine the interval for $a$, let us set a Gaussian function $G$ in the II in such a way that its maximum ($h(x) = 1$) is in $x = 0$ and its value in $x = 1$ is $r \in (0, 1)$ (see Fig. 5). In such case $b = 0$ and:

$$\exp(-(a \cdot 1 + 0)^2) = r \qquad (18)$$

From (18) we get a slope parameter for $G$:

$$a = \sqrt{-\ln(r)} = a_{\lim 1} \qquad (19)$$

Let us assume that the Gaussian AFs building the FC are not flatter than $G$. Thus, their slope parameters satisfy condition (10). Let (11) defines the steepest Gaussian AF possible. Thus, the slope parameter of the $i$-th AF can be generated from ranges (12). After substituting (19) and (18) in (12) and simplifying notation we obtain:

$$|a_i| \in \left[\sqrt{-\ln(r)}, s \cdot \sqrt{-\ln(r)}\right] \qquad (20)$$

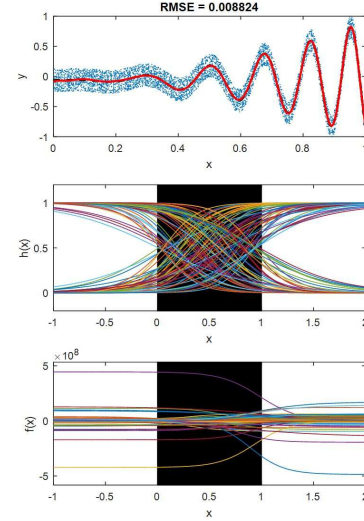

Fig. 6. Results of fitting for FNNRHN with 100 hidden sigmoid nodes, proposed algorithm with $r = 0.1$ and $s = 3$.

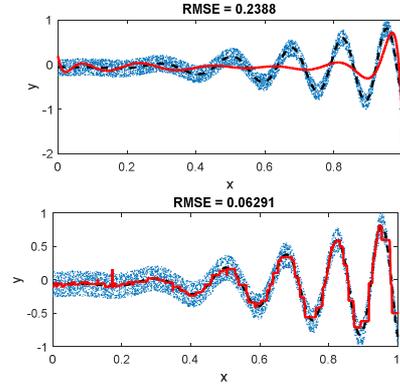

Fig. 7. Results of fitting for $s = 1.2$ (top) and $s = 10000$ (bottom).

Let us set the shift parameter $b$ in such a way that the maximum of the Gaussian AF is inside the II. So, for some $x \in [0, 1]$ we get:

$$\exp(-(a \cdot x + b)^2) = 1 \qquad (21)$$

From (21) we get the same equation for the shift parameter as for a sigmoid (15) and, consequently, the same interval from which the bias should be generated: (16).

Results of fitting when using Gaussian AFs, 100 hidden nodes and the above described approach for generating random weights and biases in Fig. 8 are shown. It was assumed: $r = 0.6$ and $s = 10$. For such value of parameters from (20) we get: $|a_i| \in [0.71, 7.15]$.

C. *Softplus AFs*

Similar considerations are performed for the softplus function:

$$h(x) = \ln(1 + \exp(a \cdot x + b)) \qquad (22)$$

Parameters $a$ and $b$ play the same role as for sigmoid and Gaussian AFs. To determine the interval for $a$, let us assume $b = 0$ (no shift). In such case for any $a$ a softplus function value in $x = 0$ is $h(0) = \ln(2)$. Now, let us assume that the value of

the softplus function $P$ in $x = 1$ is $r \in (0, \ln(2))$ (see Fig. 5). Thus:

$$\ln(1 + \exp(a \cdot 1 + 0)) = r \quad (23)$$

From (23) we get a slope parameter for $P$:

$$a = \ln(\exp(r) - 1) = a_{\lim 1} \quad (24)$$

Let us assume that AFs are not flatter than $P$. It means that their slope parameters satisfy condition (10). As in the case of sigmoid and Gaussian AFs, we assume the slope parameter for the steepest AF in the set of hidden nodes as (11). This leads to the following interval for the slope parameter of the $i$-th AF:

$$|a_i| \in \left[-\ln(\exp(r) - 1), -s \cdot \ln(\exp(r) - 1)\right] \quad (25)$$

Now, let us set a softplus function so that their most curved fragment is in the $H$. The most curved fragment is around $h(x) = \ln(2)$. So, for some $x \in [0, 1]$ we get:

$$\ln(1 + \exp(a \cdot x + b)) = \ln(2) \quad (26)$$

Again we obtain the same equation for $b$ (15) and the same interval for $i$-th AF bias (16).

In Fig. 9 results of fitting using 100 hidden nodes with softplus AFs are shown. It was assumed: $r = 0.3$ and $s = 10$. For such value of parameters from (25) we get: $|a_i| \in [1.05, 10.50]$.

### D. Sine and cosine AFs

Let us consider cosine as a AF:

$$h(x) = \cos(a \cdot x + b) \quad (27)$$

As before parameters $a$ and $b$ decide about slope and shift, respectively. For $b = 0$ and any $a$ the cosine function value in $x = 0$ is $h(0) = 1$. Let us assume that in $x = 1$ the value of the cosine function $C$ is $r \in [-1, 1)$ (see Fig. 5). Thus:

$$\cos(a \cdot 1 + 0) = r \quad (28)$$

and the slope parameter for $C$ is:

$$a = \arccos(r) = a_{\lim 1} \quad (29)$$

As before let us assume that AFs are not flatter than $C$, and not steeper than the cosine function with a slope parameter defined as (11). This leads to the following interval for $a_i$:

$$|a_i| \in [\arccos(r), s \cdot \arccos(r)] \quad (30)$$

Let us set the cosine function in the $H$ so that for some $x \in [0, 1]$:

$$\cos(a \cdot x + b) = 1 \quad (31)$$

From (31) we obtain equation for $b$ (15) and the the interval for $i$-th AF bias (16).

For sine AF exactly the same equations for weight interval (30) and biases interval (16) can be used. This is because sine function is shifted version of the cosine function.

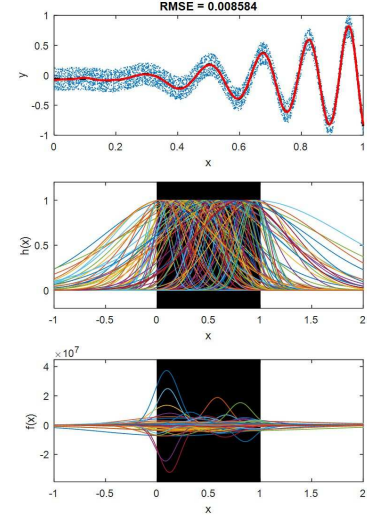

Fig. 8. Results of fitting for FNNRHN with 100 hidden Gaussian nodes, proposed algorithm with $r = 0.6$ and $s = 10$.

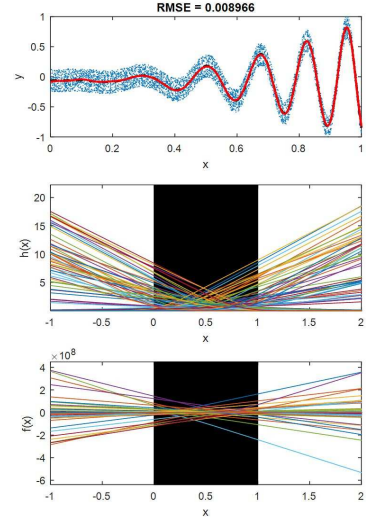

Fig. 9. Results of fitting for FNNRHN with 100 hidden softplus nodes, proposed algorithm with $r = 0.3$ and $s = 10$.

In Fig. 10 results of fitting using 100 hidden nodes with cosine AFs are shown. It was assumed: $r = 0.2$ and $s = 20$. For such value of parameters from (30) we get: $|a_i| \in [1.37, 27.38]$.

### IV. GENERATING RANDOM WEIGHTS AND BIASES - MULTIDIMENSIONAL CASE

Let input vectors $\mathbf{x}$ be normalized so that they belong to the hypercube $H = [0, 1]^n$. Similarly to the one-dimensional case, weights $a$ decide about slopes of the AF (in different directions in $n$-dimensional space) and the bias $b$ shifts AF along the x-axes. Our goal is to find a method of generating random weights and biases, to ensure that inside the hypercube $H$ there are nonlinear, steep fragments of AFs.

In this Section for brevity, we use the following acronyms:
- TF: target function $g(\mathbf{x})$,
- FS: fitted surface $\varphi(\mathbf{x})$.

To illustrate results two-variable TF is used of the form

(see Fig. 11):

$$g(\mathbf{x}) = \sin(20 \cdot \exp(x_1)) \cdot x_1^2 + \sin(20 \cdot \exp(x_2)) \cdot x_2^2 \quad (32)$$

where $x_1, x_2 \in [0, 1]$.

A variation of function (32) is the lowest at the corner [0, 0] and increases towards the corner [1, 1]. The training set contains 5000 points $(\mathbf{x}_l, y_l)$, where components of $\mathbf{x}_l$, $x_{l,1}$ and $x_{l,2}$, are independently uniformly randomly distributed on [0, 1] and $y_l$ are distorted by adding the uniform noise distributed in [–0.2, 0.2]. The testing set of the same size is created similarly but without noise. The outputs are normalized into the range [–1, 1].

### A. Sigmoid AFs

Let us set a sigmoid $S$:

$$h(\mathbf{x}) = \frac{1}{1 + \exp(-(\mathbf{a}^T \mathbf{x} + b))} \quad (33)$$

inside the hypercube $H$ in such a way that an inflection point (which is for $h(\mathbf{x}) = 0.5$) is located in the corner $\mathbf{c}_0 = [0, 0, ..., 0]$ and the sigmoid value in the opposite corner $\mathbf{c}_1 = [1, 1, ..., 1]$ is $r \in (0, 0.5)$ (see top, left chart in Fig. 12 for two-dimensional example). In such a case the shift parameter $b = 0$ and the function value in the corner $\mathbf{c}_1$ is:

$$\frac{1}{1 + \exp\left(-\left(\sum_{k=1}^{n} a_k \cdot 1 + 0\right)\right)} = r \quad (34)$$

From (34) after transformations we get:

$$\sum_{k=1}^{n} a_k = -\ln\left(\frac{1-r}{r}\right) = \Sigma_{\lim 1} \quad (35)$$

Let us assume that the AFs building the FC are not flatter in the direction $\overrightarrow{\mathbf{c}_0 \mathbf{c}_1}$ than the function $S$, and are not steeper in this direction than the sigmoid $S'$ for which the sum of the slope parameters is:

$$\sum_{k=1}^{n} a_k = \Sigma_{\lim 2} = s \cdot \Sigma_{\lim 1} \quad (36)$$

where $s > 1$.

To keep the steepness in the direction $\overrightarrow{\mathbf{c}_0 \mathbf{c}_1}$ of the $i$-th AF between the assumed boundaries the sum of its slope parameters should be from the interval:

$$\Sigma_i \in [-|\Sigma_{\lim 2}|, -|\Sigma_{\lim 1}|] \cup [|\Sigma_{\lim 1}|, |\Sigma_{\lim 2}|] \quad (37)$$

After substituting (35) and (36) in (37) the interval for $\Sigma_i$ takes the form:

$$|\Sigma_i| \in \left[\ln\left(\frac{1-r}{r}\right), s \cdot \ln\left(\frac{1-r}{r}\right)\right] \quad (38)$$

The set of weights $a_1, a_2, ..., a_n$ for a given AF is generated as follows. First, the sum $\Sigma_i$ is randomly selected from the interval (38). Then, the set of $n$ i.i.d. numbers is generated randomly: $\zeta_1, \zeta_2, ..., \zeta_n \sim U(-1, 1)$. These numbers are recalculated such that their sum is $\Sigma_i$. After recalculation we get our weights for the $i$-th AF:

$$a_{i,k} = \zeta_k \frac{\Sigma_i}{\sum_{j=1}^{n} \zeta_j} \quad (39)$$

Now, having weights $a_{i,k}$ the bias for $i$-th AF is determined in such a way that the inflection point of AF located in $\mathbf{c}_0$ for $b = 0$, is shifted to some point $\mathbf{x}$ randomly generated inside the hypercube $H$. So, for some $\mathbf{x}$: $x_1, x_2, ..., x_n \sim U(0, 1)$ we get:

$$\frac{1}{1+\exp(-(\mathbf{a}_i^T \mathbf{x} + b_i))} = 0.5 \qquad (40)$$

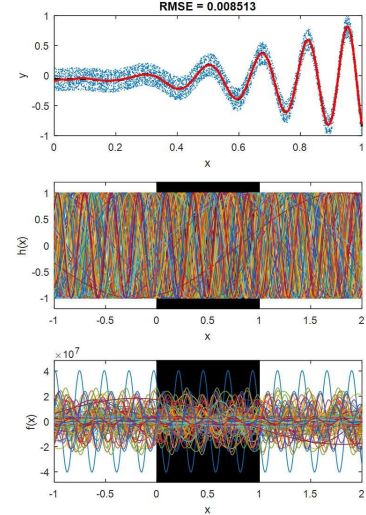

Fig. 10. Results of fitting for FNNRHN with 100 hidden cosine nodes, proposed algorithm with $r = 0.2$ and $s = 20$.

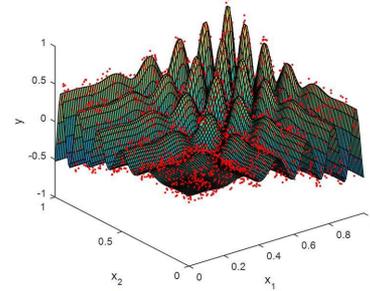

Fig. 11. Target function and training points.

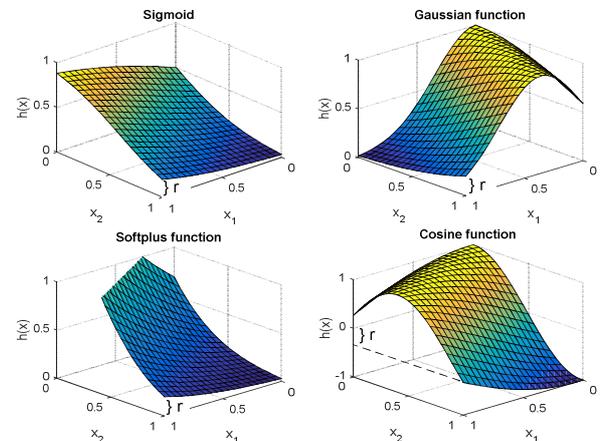

Fig. 12. Examples of AFs considered as the flattest in $H$ in the direction $\overrightarrow{\mathbf{c}_0 \mathbf{c}_1}$.

From (40) we obtain:

$$b_i = -\mathbf{a}_i^T \mathbf{x} \qquad (41)$$

Thus, the general rule for generating randomly the bias for the $i$-th AF in the case of $H = [0, 1]^n$, can be given as:

$$b_i = -\sum_{k=1}^{n} a_{i,k} x_k \qquad (42)$$

where $x_k \sim U(0, 1)$.

In Fig. 13 the FC is shown when using NNRHN with 500 sigmoid nodes. For $r = 0.1$ and $s = 5$ from (38) we get: $|\Sigma_i| \in [2.20, 10.99]$.

*B. Gaussian AFs*

Let us set a Gaussian function $G$:

$$h(\mathbf{x}) = \exp(-(\mathbf{a}^T \mathbf{x} + b)^2) \qquad (43)$$

in the hypercube $H$ in such a way that a maximum point (which is for $h(\mathbf{x}) = 1$) is located in the corner $\mathbf{c}_0 = [0, 0, ..., 0]$ and the function $G$ value in the opposite corner $\mathbf{c}_1 = [1, 1, ..., 1]$ is $r \in (0, 1)$ (see Fig. 12 for two-dimensional example). In such a case the shift parameter $b = 0$ and the function value in the corner $\mathbf{c}_1$ is:

$$\exp\left(-\left(\sum_{k=1}^{n} a_k \cdot 1 + 0\right)^2\right) = r \qquad (44)$$

From (44) we get a condition for the slope parameters of $G$:

$$\sum_{k=1}^{n} a_k = \sqrt{-\ln(r)} = \Sigma_{\lim 1} \qquad (45)$$

As for a sigmoid, let us assume that the Gaussian AFs of the hidden nodes are not flatter in the direction $\overrightarrow{\mathbf{c}_0 \mathbf{c}_1}$ than the function $G$, and are not steeper in this direction than the Gaussian function $G'$ for which the sum of the slope parameters is (36). Thus, the sum of the slope parameters of the $i$-th Gaussian AF should be from interval (37). After substituting (45) and (36) in (37) this can be written as:

$$|\Sigma_i| \in \left[\sqrt{-\ln(r)}, s \cdot \sqrt{-\ln(r)}\right] \qquad (46)$$

The weights $a_{i,1}, a_{i,2}, ..., a_{i,n}$ are generated in the same way as for sigmoid AFs from (39).

The bias of the $i$-th AF is determined in such a way that the maximum point located in $\mathbf{c}_0$ for $b = 0$ is shifted to some randomly generated point $\mathbf{x}$ inside the hypercube $H$. For this $\mathbf{x}$ we get:

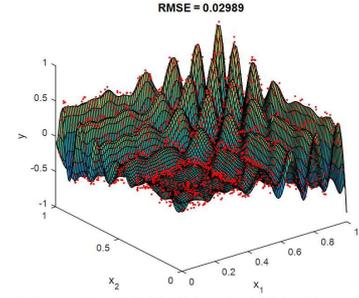

Fig. 13. Results of fitting for FNNRHN with 500 hidden sigmoid nodes, proposed algorithm with $r = 0.1$ and $s = 5$.

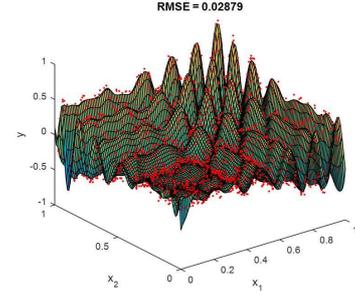

Fig. 14. Results of fitting for FNNRHN with 500 hidden Gaussian nodes, proposed algorithm with $r = 0.6$ and $s = 10$.

$$\exp(-(\mathbf{a}_i^T \mathbf{x} + b_i)^2) = 1 \qquad (47)$$

From (47) we obtain formulas for $b_i$: (41) and (42).

The FC when using FNNRHN with 500 Gaussian nodes in Fig. 14 is shown. It was assumed $r = 0.6$ and $s = 10$. For these values of parameters from (46) we get: $|\Sigma_i| \in [0.71, 7.15]$.

*C. Softplus AFs*

When in the softplus function $P$:

$$h(\mathbf{x}) = \ln(1 + \exp(\mathbf{a}_i^T \mathbf{x} + b)) \qquad (48)$$

we set $b = 0$, its value in $\mathbf{c}_0 = [0, 0, ..., 0]$ is $\ln(2)$. Let us assume that in $\mathbf{c}_1 = [1, 1, ..., 1]$ the value of $P$ is $r \in (0, \ln(2))$ (see Fig. 12). In such case:

$$\ln\left(1 + \exp\left(\sum_{k=1}^{n} a_k \cdot 1 + 0\right)\right) = r \qquad (49)$$

From (49) we get a condition for the slope parameters of $P$:

$$\sum_{k=1}^{n} a_k = \ln(\exp(r) - 1)) = \Sigma_{\lim 1} \qquad (50)$$

As for the sigmoid and Gaussian AFs let us assume that the softplus AFs are not flatter in the direction $\overrightarrow{\mathbf{c}_0 \mathbf{c}_1}$ than the function $P$, and are not steeper in this direction than the function $P'$ for which the sum of the slope parameters is (36). It means that the sum of the slope parameters for the $i$-th AF should be from interval (37). This can be written as:

$$|\Sigma_i| \in \left[-\ln(\exp(r) - 1), -s \cdot \ln(\exp(r) - 1)\right] \qquad (51)$$

The set of weights $a_{i,1}, a_{i,2}, ..., a_{i,n}$ for a given AF is generat-ed from (37).

To determine the bias $b_i$ of the softplus AF, we shift the softplus function with slopes parameters $\mathbf{a}_i$ and $b = 0$, in such a way that the point located in $\mathbf{c}_0$ is shifted to some randomly generated point $\mathbf{x}$ inside the hypercube $H$. So, for some $\mathbf{x}$: $x_1, x_2, ..., x_n \sim U(0, 1)$ we get:

$$\ln(1 + \exp(\mathbf{a}_i^T \mathbf{x} + b_i)) = \ln(2) \quad (52)$$

From (52) we obtain formulas for $b_i$: (41) and (42).

In Fig. 15 the FC is shown when using FNNRHN with 500 softplus nodes. For $r = 0.1$ and $s = 10$ from (46) we get: $|\Sigma_i| \in [2.25, 22.52]$.

*D. Sine and Cosine AFs*

The value of cosine function $C$:

$$h(\mathbf{x}) = \cos(\mathbf{a}_i^T \mathbf{x} + b) \quad (53)$$

in $\mathbf{c}_0 = [0, 0, ..., 0]$ for $b = 0$ is 1. Let us assume that in $\mathbf{c}_1 = [1, 1, ..., 1]$ the value of $C$ is $r \in [-1, 1)$ (see Fig. 12). Thus:

$$\cos\left(\sum_{k=1}^{n} a_k \cdot 1 + 0\right) = r \quad (54)$$

A condition for the slope parameters of $C$ derived from (54) is:

$$\sum_{k=1}^{n} a_k = \arccos(r) = \Sigma_{\lim 1} \quad (55)$$

Let us assume that the cosine AFs are not flatter in the direction $\overrightarrow{\mathbf{c}_0 \mathbf{c}_1}$ than the function $C$, and are not steeper in this direction than the function $C'$ for which the sum of the slope parameters is (36). Thus, the sum of the slope parameters for the $i$-th cosine AF should be from interval (37), which can be written as:

$$|\Sigma_i| \in [\arccos(r), s \cdot \arccos(r)] \quad (56)$$

Having the sum $\Sigma_i$, the set of weights $a_{i,1}, a_{i,2}, ..., a_{i,n}$ for the $i$-th cosine AF is generated from (37).

To determine the bias $b_i$ of the softplus AF, we shift the cosine function with slopes parameters $\mathbf{a}_i$ and $b = 0$, in such a way that the point located in $\mathbf{c}_0$ is shifted to some randomly generated point $\mathbf{x} \sim U(0, 1)^n$ inside the hypercube $H$. Thus:

$$\cos(\mathbf{a}_i^T \mathbf{x} + b_i) = 1 \quad (57)$$

From (57) we obtain formulas for $b_i$: (41) and (42).

The same intervals for weights (56) and biases (42) can be assumed for sine AFs.

In Fig. 16 the FC is shown when using NNRHN with 500 cosine nodes. For $r = 0.2$ and $s = 50$ from (56) we get: $|\Sigma_i| \in [1.37, 68.47]$.

In the above analysis we quietly assumed that input points are evenly distributed in a hypercube $H$. In many (or even mostly) cases they are not. In such a case it is reasonably to shift the AFs in the bias determination step from $\mathbf{c}_0$ not to some randomly selected point $\mathbf{x}$ but to one of the training point. This ensures that all AFs have their nonlinear fragments in the regions containing data. The only modification of the above method is that for generating biases in (42) we use: $[x_1, x_2, ..., x_n] = \mathbf{x}_\xi \in \Phi$, where $\xi$ is a random integer uniformly distributed between 1 and $N$. Alternative way of choosing $\mathbf{x}_\xi$ is to select them in regions of the input space where the TF is the most variable (has steep fragments). Another idea to calculate biases $b_i$ is to group training points into $m$ clusters. The prototypes $\mathbf{p}$ of these clusters (e.g. centroids) can be taken as the points to which the AFs are shifted from $\mathbf{c}_0$.

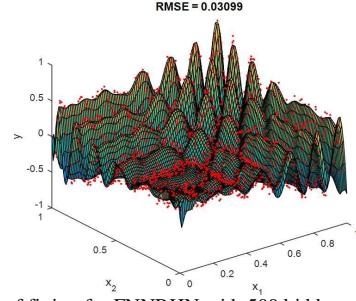

Fig. 15. Results of fitting for FNNRHN with 500 hidden softplus nodes, proposed algorithm with $r = 0.1$ and $s = 10$.

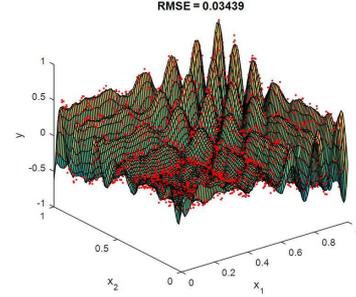

Fig. 16. Results of fitting for FNNRHN with 500 hidden cosine nodes, proposed algorithm with $r = 0.2$ and $s = 50$.

The above analysis for multidimensional case in Table I are summarized. The proposed process of generating random weights and biases for FNNRHN is shown in Algorithm 1. It requires the inputs to be normalized: $\mathbf{x} \in H = [0, 1]^n$.

V. SIMULATION STUDY

In this section the proposed method of FNNRHN random parameters generation are illustrated on several examples. Results are compared with the state-of-the-art method proposed recently in [16] as well as with Modified Quickprop [21] and Incremental Random Vector Functional Link (IRVFL) network [10]. Results of the comparative models are taken from [16] as well as regression problems including a function approximation and three real-world modeling tasks:

- Approximation of the single-variable TF:

$$g(\mathbf{x}) = 0.2e^{-(10x-4)^2} + 0.5e^{-(80x-40)^2} + 0.3e^{-(80x-20)^2} \quad (58)$$

The training set contains 1000 points $(x_l, y_l)$, where $x_l$ are uniformly randomly distributed on $[0, 1]$. The test set of size 300 is generated from a regularly spaced grid on $[0,1]$.

- Stock - daily stock prices from January 1988 through October 1991, for ten aerospace companies. The task is to approximate the price of the 10th company given the prices of the rest. There are 950 samples composed of nine input variables and one output variable. The whole data set was divided into training set containing 75% samples selected randomly, and the test set containing the remaining samples.
- Concrete - the dataset contains the concrete compressive strength, age, and ingredients: cement, blast furnace slag, fly ash, water, superplasticizer, coarse aggregate, and fine aggregate. The task is to approximate the highly nonlinear relationship between concrete compressive strength and the ingredients and age. There are 8192 samples composed of eight input variables and one output variable. The whole data set was divided into training and test parts in the same manner as Stock data set.
- Compactiv - the Computer Activity dataset is a collection of computer systems activity measures. The data was collected from a Sun Sparcstation 20/712 with 128 Mbytes of memory running in a multi-user university department. The task is to predict the portion of time that CPUs run in user mode. There are 8192 samples composed of 21 input variables (activity measures) and one output variable. The whole data set was divided into training and test parts in the same manner as Stock data set.

The datasets Stock, Concrete and Compactiv were downloaded from KEEL (Knowledge Extraction based on Evolutionary Learning) dataset repository (http://www.keel.es/). The input and output variables are normalized into [0,1]. All results reported in this work take averages over 100 independent trials. Root Mean Squares Error (RMSE) was used as a measure of modeling accuracy.

The comparative models adopted from [16] are:
- MQ - Modified Quickprop algorithm proposed in [21] that iteratively finds the appropriate parameters for the new hidden node added in the incremental procedure. The parameters of MQ were set by authors [16] as follows: learning rate = 0.05, maximum iterative number = 200.
- IRVFL - Incremental Random Vector Functional Link network where the model is built incrementally with random assignment of the input weights and biases, and constructive evaluation of its output weights using the least squares method [10]. The random parameters were taken by authors of [16] from the uniform distribution over [−1, 1].
- SCN - Stochastic Configuration Network proposed in [16]. This is a variant of IRVFL with random parameters generated with an inequality constraint from the adaptively selected scope [−$\lambda$, $\lambda$], ensuring the universal approximation property of the built randomized learner model. Among three algorithmic implementations of SCN, the most accurate one was chosen, signed SC-III in [16], where the output weights are recalculated all together through solving a global least squares problem each time a new hidden node is added. Sigmoidal activation function were used for the hidden nodes. The SCN parameters were selected by authors of [16] to ensure the best performance.

Table II shows the results: errors and their standard deviations for FNNRHN with different AFs which parameters are generated using the proposed method (FNNRHN-sig, FNNRHN-Gauss, FNNRHN-cos and FNNRHN-soft) as well as for the comparative models (copied from table I and II of [16]). The optimal parameter values of the proposed method are also shown in Table II. They are selected in the grid search using 10-fold cross-validation. In these procedure only the training parts of the datasets were used. The number of hidden neurons for FNNRHN was set to 100 in all cases.

As we can see from Table II the proposed method allows FNNRHN to achieve results not worse than the most sophisticated comparative model SCN and outperforms MQ and IRVFL in terms of both learning and generalization. In approximation of TF (58), which is highly nonlinear having two spikes, our method shows its power and achieves significantly better results than other models. The FCs for this case are shown in Fig. 17 (compare with Fig. 3 in [16], where FC for IRVFL and SCN are shown).

Worse performance for the MQ algorithm results from a method of computing weights for new nodes added in the hidden layer. For this purpose the gradient-ascent algorithm is applied which uses also second-order information in optimiza-

TABLE I
GENERATION OF THE FNNRHN HIDDEN NODE PARAMETERS

| Activation function | Condition for the sum of input weights $\Sigma_i$ | Interval for $r$ | Weights of $i$-th hidden node | Bias of $i$-th hidden node |
|---|---|---|---|---|
| $\frac{1}{1+\exp(-(\mathbf{a}^T\mathbf{x}+b))}$ | $\lvert \Sigma_i \rvert \in \left[\ln\left(\frac{1-r}{r}\right), s \cdot \ln\left(\frac{1-r}{r}\right)\right]$ | $r \in (0, 0.5)$ | | |
| $\exp(-(\mathbf{a}^T\mathbf{x}+b)^2)$ | $\lvert \Sigma_i \rvert \in \left[\sqrt{-\ln(r)}, s \cdot \sqrt{-\ln(r)}\right]$ | $r \in (0, 1)$ | $a_{i,k} = \zeta_k \dfrac{\Sigma_i}{\sum_{j=1}^{n}\zeta_j}$ | $b_i = -\sum_{k=1}^{n} a_{i,k} x_k$ |
| $\ln(1+\exp(\mathbf{a}^T\mathbf{x}+b))$ | $\lvert \Sigma_i \rvert \in \left[-\ln(\exp(r)-1), -s \cdot \ln(\exp(r)-1)\right]$ | $r \in (0, \ln(2))$ | | |
| $\cos(\mathbf{a}^T\mathbf{x}+b)$ $\sin(\mathbf{a}^T\mathbf{x}+b)$ | $\lvert \Sigma_i \rvert \in \left[\arccos(r), s \cdot \arccos(r)\right]$ | $r \in [-1, 1)$ | | |

where: $\mathbf{x} \in [0, 1]^n$; $s > 1$; $k = 1, 2, ..., n$; $\zeta_1, \zeta_2, ..., \zeta_n$ are i.i.d $U(-1, 1)$ random variables; $x_k \sim U(0, 1)$ or $x_k = x_{\xi,k}$, where $\mathbf{x}_\xi \in \Phi$, $\xi \sim U\{1, 2, ..., N\}$ or $x_k = p_{i,k}$, where $\mathbf{p}_i$ is a prototype of the $i$-th cluster of $\mathbf{x} \in \Phi$

**Algorithm 1** Generation of the hidden node weights and biases for FNNRHN

**Input**:
    Activation function $g(\mathbf{x})$
    Number of hidden nodes $m$
    Number of inputs $n$
    Steepness parameter $s > 1$

**Output**:
$$\text{Weights } \mathbf{A} = \begin{bmatrix} a_{1,1} & \cdots & a_{m,1} \\ \vdots & \vdots & \vdots \\ a_{1,n} & \cdots & a_{m,n} \end{bmatrix}$$
    Biases $\mathbf{b} = [b_1, b_2, ..., b_m]$

**Procedure**:
1. Set $r_{\min} = \min_{\mathbf{x} \in R^n} h(\mathbf{x})$
2. Set $r_{\max} = h(\mathbf{x})$ for $\mathbf{x} = \mathbf{c}_0 = [0, 0, ..., 0]$, $b = 0$
3. Choose $r$ from $(r_{\min}, r_{\max})$
4. Transform $g(\mathbf{x})$ assuming $\mathbf{x} = \mathbf{c}_0 = [1, 1, ..., 1]$ and $b = 0$ to get formula for $\sum_{k=1}^{n} a_k$
5. Assume $\Sigma_{\lim 1} = \sum_{k=1}^{n} a_k$ and $\Sigma_{\lim 2} = s \cdot \Sigma_{\lim 1}$
    **for** each node $i = 1, 2, ..., m$ **do**
6.     Choose randomly $\Sigma_i$ from $[-|\Sigma_{\lim 2}|, -|\Sigma_{\lim 1}|] \cup [|\Sigma_{\lim 1}|, |\Sigma_{\lim 2}|]$
7.     Choose randomly i.i.d. $\zeta_1, \zeta_2, ..., \zeta_n \sim U(-1, 1)$
    **for** $k = 1, 2, ..., n$ **do**
8.     Calculate $a_{i,k} = \zeta_k \dfrac{\Sigma_i}{\sum_{j=1}^{n} \zeta_j}$
    **end for**
9a. Choose randomly i.i.d $x_1, x_2, ..., x_n \sim U(0, 1)$
9b. or set $[x_1, x_2, ..., x_n] = \mathbf{x}_\xi \in \Phi$, where $\xi \sim U\{1, 2, ..., N\}$
9c. or set $[x_1, x_2, ..., x_n] = \mathbf{p}_i$, where $\mathbf{p}_i$ is a prototype of the $i$-th cluster of $\mathbf{x} \in \Phi$
10. Calculate $b_i = -\sum_{k=1}^{n} a_{i,k} x_k$
    **end for**
11. Return $\mathbf{A}, \mathbf{b}$

TABLE II
RESULTS COMPARISON AMONG PROPOSED AND COMPARATIVE MODELS[a]

| Algorithm | Parameters $r, s$ | Training RMSE | Test RMSE |
|---|---|---|---|
| Function (58) | | | |
| MQ | | 0.1030±0.0001 | 0.1011±0.0003 |
| IRVFL | | 0.1626±0.0005 | 0.1617±0.0008 |
| SCN | | 0.0097±0.0036 | 0.0100±0.0033 |
| FNNRHN-sig | 0.04, 40 | 0.0040±0.0032 | 0.0043±0.0032 |
| FNNRHN-Gauss | 0.54, 100 | 0.0031±0.0020 | 0.0052±0.0040 |
| FNNRHN-cos | 0.08, 190 | 0.0063±0.0197 | 0.0071±0.0220 |
| FNNRHN-soft | 0.32, 120 | 0.0038±0.0026 | 0.0049±0.0038 |
| Stock | | | |
| MQ | | 0.0410±0.0014 | 0.0407±0.0017 |
| IRVFL | | 0.1853±0.0248 | 0.1787±0.0237 |
| SCN | | 0.0327±0.0007 | 0.0347±0.0012 |
| FNNRHN-sig | 0.22, 6.4 | 0.0287±0.0011 | 0.0325±0.0091 |
| FNNRHN-Gauss | 0.94, 3.6 | 0.0281±0.0010 | 0.0345±0.0095 |
| FNNRHN-cos | 0.74, 8.8 | 0.0290±0.0011 | 0.0328±0.0014 |
| FNNRHN-soft | 0.46, 3.4 | 0.0277±0.0009 | 0.0353±0.0015 |
| Concrete | | | |
| MQ | | 0.0910±0.0014 | 0.0869±0.0021 |
| IRVFL | | 0.1929±0.0135 | 0.1983±0.0166 |
| SCN | | 0.0835±0.0012 | 0.0850±0.0025 |
| FNNRHN-sig | 0.44, 2.9 | 0.0740±0.0022 | 0.0871±0.0055 |
| FNNRHN-Gauss | 0.97, 3.4 | 0.0755±0.0022 | 0.0876±0.0047 |
| FNNRHN-cos | 0.95, 1.7 | 0.0742±0.0023 | 0.0877±0.0047 |
| FNNRHN-soft | 0.52, 2.4 | 0.0737±0.0023 | 0.0882±0.0043 |
| Compactiv | | | |
| MQ | | 0.0600±0.0071 | 0.0624±0.0075 |
| IRVFL | | 0.1924±0.0283 | 0.1882±0.0281 |
| SCN | | 0.0394±0.0016 | 0.0418±0.0021 |
| FNNRHN-sig | 0.30, 1.4 | 0.0398±0.0022 | 0.0409±0.0027 |
| FNNRHN-Gauss | 0.96, 2.4 | 0.0372±0.0020 | 0.0411±0.0038 |
| FNNRHN-cos | 0.98, 2.2 | 0.0361±0.0019 | 0.0399±0.0028 |
| FNNRHN-soft | 0.66, 3.6 | 0.0358±0.0018 | 0.0434±0.0044 |

[a] Results for MQ, IRVFL and SCN are taken from [16].

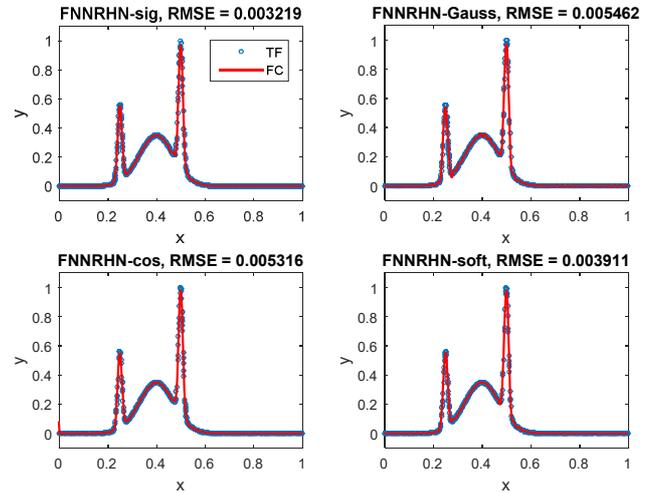

Fig. 17. Results of TF (58) fitting for FNNRHN with different AFs.

tion of the objective function. But it is problematic when one is exploring in the region of a plateau in the error surface, where the first and second derivatives of the function to be optimized with respect to all the parameters are nearly zero. Although the MQ algorithm is equipped with an escape mechanism from these regions, it does not always work. Thus, the optimal solution cannot be guaranteed when the optimization is non-convex, i.e. it is nonlinear in the hidden layer parameters.

In the case of the randomized algorithms for training NNs the optimization problem is linear in the parameters, thus the optimization is convex and has an analytic solution, such as the least squares. But in these algorithms the key issue is to properly generate the random parameters of the hidden neurons to find the orthogonal projection of $y$ into the input space [1]. In IRVFL the random parameters are taken from the fixed range [−1, 1], and there is no guarantee that it is appropriate for the regression problem. So, the results for IRVFL are even worse than for MQ. SCN searches for the random parameter ranges for each new node added to the hidden layer. Thus, this ranges are optimized for each neuron. This translates into much better results than for fixed ranges, which are set without any scientific justification. But in the light of the considerations carried out in this work, assigning the same ranges for weights and biases is questionable. There is no problem when the TF is not strongly nonlinear or "flat", without spikes and sudden jumps. Such function can be approximated using flat fragments of the AFs, so the ranges for random parameters are not as important. But the problem arises when we approximate a strongly nonlinear function.

Our proposed method generates random parameters in such a way that the most nonlinear and steepest fragments of AFs are inside the region with data. This allows the model to approximate strongly nonlinear functions, such as function (58) with spikes. But the weights corresponding to the slope parameters of the AFs are generated from different ranges than biases corresponding to the shift parameters, due to different meaning of these parameters. The mechanism of selecting random weights and biases is very simple and needs to tune only two parameters, $r$ and $s$, which control bias-variance tradeoff of the network. To select these parameters we used cross-validation. Competitive algorithm SCN works in incremental mode, which is more time consuming, and needs five parameters to be tuned ($r$, $\lambda$, $T_{\max}$, $\epsilon$, and $L$, see [16] for details). This can make this algorithm difficult to apply in practice.

## VI. Conclusion

In this work we demonstrate that the intervals of the random weights and biases in FNNRHN are extremely important due to approximation properties of the network. Activation functions of the hidden neurons are the basis functions which linear combination forms the surface fitting data. For nonlinear target function the set of AFs should deliver nonlinear fragments to model the target function in its nonlinear regions with required accuracy.

The main contribution of this work is to propose a practical method of randomly generating weights and biases in FNNRHN to set nonlinear fragments of AFs in the input space region containing data points. The analyzes carried out lead to the conclusion that parameters of hidden nodes are dependent on the input data range and activation function type. Ranges for weights and biases should be considered separately, because this parameters have different meaning. Moreover, the range for the bias of the $i$-th hidden node is strictly dependent on the weights of this node. The proposed method allows us to control the flatness and steepness of the AF set and hence the degree of generalization of the network.

The experimental results demonstrate that our approach is very promising. It shows remarkable improvement in accuracy compared with existing methods such as Modified Quickprop and incremental RVFL. It is also competitive with the latest solutions, such as Stochastic Configuration Network, having less complex algorithm.